\documentclass[journal]{IEEEtran}

\usepackage{amsmath,amssymb}
\usepackage{graphicx}
\usepackage{xcolor}
\graphicspath{{./}}
\newcommand{\includefigure}[2][width=\linewidth]{%
  \IfFileExists{#2}{\includegraphics[#1]{#2}}{%
    \fbox{\parbox{0.8\linewidth}{\centering\vspace{2em}%
    \textbf{Figure pending}\vspace{2em}}}%
  }%
}
\usepackage{cite}
\usepackage{booktabs}
\usepackage{multirow}
\usepackage{textcomp}  
\providecommand{\degree}{\ensuremath{^\circ}}

\title{Calibrating Attribution Proxies for Reward Allocation in Participatory Weather Sensing}

\author{Mark~C.~Ballandies,~%
Michael~T.~C.~Chiu,~%
and~Claudio~J.~Tessone%
\thanks{This work has been submitted to the IEEE for possible publication.
Copyright may be transferred without notice, after which this version may
no longer be accessible.}%
\thanks{M.~C.~Ballandies and C.~J.~Tessone are with the University of Zurich, Zurich, Switzerland.}%
\thanks{M.~C.~Ballandies and M.~T.~C.~Chiu are with WiHi, Zug, Switzerland.}%
}

\begin{document}

\markboth{IEEE Internet of Things Journal}{Ballandies \MakeLowercase{\textit{et al.}}: Attribution Proxies for Participatory Weather Sensing}

\maketitle

\begin{abstract}
Large-scale  IoT weather sensing networks require incentive mechanisms
to sustain participation, yet determining how much value individual
data contributions bring to the network remains an open problem.
Existing approaches address data quality but not data valuation;
in operational meteorology, adjoint-based methods derive value from
the forecast model itself but require full data assimilation
infrastructure.
We propose to utilise differentiable AI weather models to fill this
gap and characterise gradient-based attribution on gridded GFS analysis
inputs as a candidate value signal, evaluating fidelity, calibration,
cost, and gaming vulnerability across more than 400~configurations.
Attribution captures near-optimal sensor placement utility with
monotonically faithful payments, but can be inflated by adversarial inputs,
with detection requiring external baseline data.
These findings establish gradient attribution as a computationally
validated signal for model-informed reward allocation in participatory
weather sensing.
\end{abstract}


\section{Introduction}
\label{sec:intro}

Large-scale IoT sensing networks for weather and environmental monitoring
are expanding rapidly, from national citizen-science initiatives \cite{mahajan2021yourself} to
community-operated and crowdsensed
platforms~\cite{chapman2017crowdsourcing,meier2017crowdsourcing,coney2022useful}.
Sustaining active user participation is crucial for participatory sensing success \cite{heiskala2016crowdsensing,karim2020big}. 
Incentive mechanism design for crowdsensing (Stackelberg games, reverse auctions, market-based pricing~\cite{yang2012crowdsourcing,zhang2016incentives,restuccia2016incentive}) and cryptoeconomic token mechanisms for Decentralized Physical Infrastructure Networks (DePIN)~\cite{chiu2025depin,ballandies2022incentivize} all assume a value signal as input.
However, determining how much value an individual data contribution
brings to the network remains an open problem. In particular, quality assessment
addresses whether data are reliable enough to
use~\cite{chapman2017crowdsourcing,devos2019quality,fenner2021crowdqc},
but not what reliable data is worth.
Market-based data pricing~\cite{pandey2024strategic} presupposes a
data marketplace that rarely exists in participatory sensing, while
Shapley-based fair division~\cite{byun2009fair,ghorbani2019data}
requires evaluating exponentially many data subsets, making it
infeasible for real-time deployment.
In operational meteorology, adjoint-based observation impact
assessment~\cite{cardinali2009monitoring,langland2004estimation}
addresses this by deriving value from the numerical weather prediction
model itself, but requires full data assimilation infrastructure
unavailable to IoT sensing networks.
In practice, deployed networks lack such valuation:
WeatherXM~\cite{weatherxm2024rewards}, a community-operated
network of thousands of weather stations distributing daily 
rewards, allocates payments proportional to data quality and uptime
rather than downstream utility.  Consequently, the mechanism cannot
dynamically direct network growth toward locations and variables
which are most useful for the network.
We propose to utilise differentiable AI weather models to derive
an informativeness signal computable at forecast time without verification data.
The signal measures model sensitivity, how much each input location
and variable moves the forecast, rather than demonstrated loss
reduction; Sections~\ref{sec:results_validity}--\ref{sec:results_incentive}
validate that this sensitivity signal nevertheless yields near-optimal
sensor placement and calibrated payments.

Specifically, we characterise gradient-based attribution in
differentiable AI weather models (FourCastNet and SFNO) as a candidate
value signal, evaluating its fidelity against ablation-based reference
utility across more than 400~configurations and its gaming resilience
across 3{,}700 attack scenarios, evaluating on gridded GFS analysis
inputs under controlled conditions (the station-to-grid link is the
priority extension; Section~\ref{sec:discussion}).
The main contributions are:
\begin{enumerate}
  \item Attribution ranks input variables faithfully, with a regime-conditional
    spatial advantage in complex terrain and wind forecasting
    (+1.4~pp FCN, 73\% win rate).
  \item The cheapest single-pass variant (Gradient~$\times$~Input; GTI) retains 83\% of Integrated Gradients (IG)
    fidelity at $1/50$th the computational cost.
  \item Near-optimal sensor placement ($\geq 92\%$ oracle utility)
    with monotonically calibrated payments (overpayment 33--36\%
    vs.\ 47--55\% for distance, 61--72\% for uniform) and
    payment shares stable to within 15--23\% across forecast cycles.
  \item Anomaly inflation is detectable with baseline
    proxy-monitoring (100\% top-5 hit rate, SFNO) but baseline-free
    detection fails entirely; data-fabrication spoofing evades
    detection, requiring complementary staking and identity mechanisms (Section~\ref{sec:discussion_deployment}).
\end{enumerate}

\section{Related Work}
\label{sec:related}

\subsection{Participatory Weather Sensing and Incentive Design}
\label{sec:related_quality}

Citizen weather station networks have been validated for urban climate
monitoring~\cite{chapman2017crowdsourcing,meier2017crowdsourcing,coney2022useful},
with quality control advancing from spatial-consistency
filtering~\cite{devos2019quality} to reference-free statistical
methods~\cite{fenner2021crowdqc}.
WeatherXM~\cite{weatherxm2024rewards,weatherxm2024qod} is a
token-incentivised weather station network (thousands of stations,
80+~countries) whose rewards are proportional to data quality and
availability, not predictive contribution; Sybil resistance relies on
mandatory proprietary hardware (${\sim}$\$300--500), which raises
attack cost but restricts participation to approved
devices~\cite{chiu2025depin}.
To our knowledge, no deployed participatory sensing network implements
model-derived valuation; existing reward allocation is geometry- or
quality-based.
The informativeness signal we characterise addresses this gap.

\subsection{Data Valuation and Observation Impact}
\label{sec:related_valuation}

\paragraph{Shapley-based data valuation.}
Shapley-based methods~\cite{ghorbani2019data,jia2019towards,kwon2022beta,byun2009fair}
quantify a datum's marginal contribution across all coalitions and satisfy
axioms of fair division, but evaluating the $O(2^N)$ coalitions is
infeasible for real-time reward allocation.
Integrated Gradients~(IG)~\cite{sundararajan2017axiomatic} satisfies a
completeness axiom analogous to Shapley efficiency at $O(K)$ backward
passes, providing a tractable gradient-based alternative.

\paragraph{Observation impact in meteorology.}
Adjoint-based Forecast Sensitivity to Observation Impact
(FSOI)~\cite{langland2004estimation,gelaro2010thorpex,cardinali2009monitoring}
quantifies per-observation forecast impact operationally, and the
socioeconomic value of surface observations is well
documented~\cite{kull2021value}.
However, these approaches require full data assimilation infrastructure
unavailable to IoT sensing networks.
AI weather models (FourCastNet~\cite{pathak2023fourcastnet},
SFNO~\cite{bonev2023spherical}, GraphCast~\cite{lam2023learning},
Pangu-Weather~\cite{bi2023accurate}, Aurora~\cite{bodnar2025aurora})
are differentiable end-to-end, enabling gradient-based attribution
\emph{at inference time} without that infrastructure.
Prior work demonstrates that gradient attribution captures physically
meaningful patterns in weather and climate
models~\cite{mcgovern2019making,toms2020physically,banomedina2025},
but has not evaluated whether such attribution can serve as a
calibrated, gaming-resilient value signal for participatory sensing
reward allocation, which are the operational requirements that mechanism
deployment demands.

\section{Attribution Signal and Evaluation Design}
\label{sec:methods}

\subsection{Problem Formulation and Payment Rule}
\label{sec:incentive}

We frame gradient-based attribution as a candidate informativeness signal
for participatory sensing reward mechanisms and evaluate it via sensor
selection and proportional payment allocation.  Given a budget of
$K$ sensors (out of $N = 468$ European grid points), we compare six
selection strategies, three of which use gradient-based attribution
scores (defined in Section~\ref{sec:ig}):
\begin{enumerate}
  \item \textbf{IG-based}: select the $K$ locations with the highest
    time-averaged IG importance $\sum_v |\mathrm{IG}_{v,g}|$.
  \item \textbf{GTI-based}: select the $K$ locations with the highest
    time-averaged Gradient~$\times$~Input importance.
  \item \textbf{VG-based}: select the $K$ locations with the highest
    time-averaged Vanilla Gradient importance.
  \item \textbf{Distance-based}: select the $K$ locations closest
    to the forecast target (haversine distance).
  \item \textbf{Uniform}: random selection (expected utility $=K/N$).
  \item \textbf{Oracle}: select the $K$ locations with the highest ablation
    utility $|U_g^{\text{spatial}}|$ (upper bound; requires reference utility).
\end{enumerate}
Captured utility is the fraction of total ablation utility captured by
the selected set $S \subseteq \{1,\dots,N\}$ of $K$~stations, where
$U_g$ is the ablation utility at station~$g$ (defined in
Section~\ref{sec:validation}):
\begin{equation}
  C(S) = \frac{\sum_{g \in S} |U_g|}{\sum_{g=1}^{N} |U_g|}
  \label{eq:captured_utility}
\end{equation}
We report the efficiency ratio $C_{\text{method}} / C_{\text{uniform}}$ and the
optimality ratio $C_{\text{method}} / C_{\text{oracle}}$.
Eq.~\eqref{eq:captured_utility} treats station contributions as
additive; subadditivity arising from spatial redundancy is examined
in Section~\ref{sec:discussion}.

For reward allocation, each sensor receives payment proportional to
attribution magnitude:
\begin{equation}
  p(g) = \frac{|A(g)|}{\sum_{g'} |A(g')|} \cdot B
  \label{eq:payment}
\end{equation}
where $A(g)$ is the importance score assigned to station~$g$ (instantiated
as IG, GTI, or VG attribution (Section~\ref{sec:ig}), inverse haversine
distance, or uniform weight, depending on the method under
evaluation) and $B$ is the total budget.
This rule is budget-balanced by construction ($\sum_g p(g) = B$) and
individually rational ($p(g) \geq 0$).

\paragraph{Prediction vs.\ loss attribution.}
As noted in Section~\ref{sec:intro}, the attribution proxy
differentiates the model prediction~$F(x)$ at the target location
rather than the forecast loss $|F(x) - y^*|$, because verification
data~$y^*$ is unavailable when real-time rewards are computed.
Sections~\ref{sec:results_validity}--\ref{sec:results_incentive}
validate that this prediction-sensitivity signal correlates with
loss-based ablation utility and yields near-optimal sensor placement,
confirming operational usefulness despite the objective gap.

\paragraph{Proxy-utility calibration and overpayment.}
A proxy is \emph{calibrated} if stations binned by proxy score exhibit
monotonically increasing mean ablation utility (distinct from
probabilistic calibration~\cite{dawid1982well}).
\emph{Overpayment} at station $g$ is
$\max(0,\, p_{\text{proxy}}(g) - p_{\text{true}}(g))$, where
$p_{\text{true}}(g) = |U_g| / \sum_{g'} |U_{g'}|$; total overpayment
bounds budget misallocation.  By budget balance, total underpayment equals
total overpayment.
We report Gini ratios and per-station overpayment scatter in
Section~\ref{sec:results_calibration}; definitions are detailed in
Appendix~\ref{sec:appendix_setup_methods}.

\paragraph{Assumptions.}
Stations report to a central coordinator with fixed budget~$B$ per
forecast window.  Ablation audits require verification data and are feasible offline
but not at forecast time; in deployment, they serve as periodic
calibration checks on the proxy signal rather than the real-time
allocation rule.  Attackers inflate anomalies at their own station(s);
collusion and adversarial input crafting are not modelled; these
represent distinct multi-agent and model-targeted threat models
outside the scope of single-station gaming analysis.
Full assumption detail is in
Appendix~\ref{sec:appendix_setup_methods}.

\subsection{Attribution Proxy Family}
\label{sec:ig}

Integrated Gradients (IG)~\cite{sundararajan2017axiomatic} attributes
target predictions by integrating gradients from a baseline $\bar{x}$
to the input $x$:
\begin{equation}
  \mathrm{IG}_i = (x_i - \bar{x}_i) \cdot \frac{1}{K}
  \sum_{k=0}^{K} w_k \,
  \frac{\partial F}{\partial x_i}\bigg|_{\bar{x} + \frac{k}{K}(x - \bar{x})}
  \label{eq:ig}
\end{equation}
where $K=50$ and $\bar{x}$ is the per-variable climatological mean
(trapezoidal quadrature: $w_0 = w_K = 0.5$, $w_k = 1$ otherwise).
The scalar index $i$ decomposes as $(v, \ell)$ for variable~$v$ at
spatial location~$\ell$; we write $A_{v,\ell}$ for the attribution
score of any proxy in the family.
Two cheaper alternatives complete the proxy family:
\textbf{Gradient $\times$ Input (GTI)}, $\mathrm{GTI}_i =
(x_i - \bar{x}_i) \cdot \partial F / \partial x_i$ (single backward pass,
first-order IG approximation); and
\textbf{Vanilla Gradients (VG)}~\cite{simonyan2014deep}, $\mathrm{VG}_i = \partial F / \partial x_i$
(single pass, no input-relative scaling).
A model-free \textbf{Distance} baseline (inverse haversine) completes the
comparison.  Table~\ref{tab:proxy_properties} summarises cost, axiomatic
properties, and empirical performance.
\emph{Ablation} serves as the offline audit (requires verification data; periodic); \emph{proxy
scores} power the online allocation rule (computable from the prediction alone; every forecast cycle; cost independent of station count).

\subsection{Weather Models}
\label{sec:models}

We evaluate two deterministic AI weather models that produce 6-hour forecasts
from Global Forecast System (GFS) initial conditions.

\paragraph{FourCastNet (FCN).} Vision-transformer model~\cite{pathak2023fourcastnet}
on a $720 \times 1440$ grid at $0.25\degree$ resolution with 26 input variables.

\paragraph{Spherical Fourier Neural Operator (SFNO).} Spectral
model~\cite{bonev2023spherical} on the $721 \times 1440$ native GFS grid
with 73 input variables.  FCN and SFNO share 24 common variables; all
model-comparable attribution and ablation analyses are restricted to this
common set.

\paragraph{Payment uncertainty.}
\label{sec:payment_uncertainty}
We quantify payment stability via bootstrap 95\% CIs
on per-station payment shares (resampling individual timestamps with
replacement).  An optional shrinkage estimator blends
the proxy with a distance prior (details in
Appendix~\ref{sec:appendix_setup_methods}).

\subsection{Gaming Resilience Assessment}
\label{sec:gaming_methods}

We evaluate gaming resilience via anomaly-inflation simulations.
We systematically vary three threat dimensions, attack magnitude,
attacker proximity, and variable scope, to characterise the
attribution signal's attack surface.  In each scenario, $n$ malicious stations ($n \in \{1, 3, 5\}$) inflate their
local anomaly by a factor $(1 + \text{pct}/100)$ for a subset of
variables (t2m only, u10m only, or all surface variables).  The base
design uses 3~magnitudes (10/30/50\%) $\times$ 10~random seeds with
uniform random attacker placement from the 468-station European grid,
yielding 270 scenarios per model--city--variable
configuration.  For the Zurich/t2m target, we extend the design with
stratified attacker placement (close $<500$\,km, mid 500--1500\,km,
mixed) and additional magnitudes (100\%, 200\%), yielding 1{,}230
inflation scenarios per model, plus 80 climatological-mean spoofing
scenarios per model in which malicious stations replace their input
with the long-term mean (a zero-cost data-fabrication attack).  The
combined experiment totals 3{,}700 scenarios across six
configurations.

We evaluate detectors from three families
(full taxonomy in the supplementary material, Section~S8):
baseline-dependent detectors comparing gaming-period scores against
a pre-attack baseline,
a supervised classifier (logistic regression and gradient boosting with
leave-one-configuration-out CV),
and baseline-free detectors using only single-snapshot statistics.
Gaming detection uses \emph{signed} attributions, whereas reward
allocation uses \emph{unsigned} $|A(g)|$; complementary integrity
mechanisms such as staking, where participants lock economic deposits
that can be slashed for misbehaviour~\cite{chiu2025depin}, are not
evaluated here.
Detection quality is measured via top-$k$ hit rate and PR-AUC;
per-configuration results appear in Table~\ref{tab:gaming_summary}.

\subsection{Validation Protocol and Experimental Setup}
\label{sec:validation}

\paragraph{Global validation.}
\label{sec:global_ablation}
For each of the 24 common input variables, we replace the entire field with its
climatological mean, run the model forward, and measure the change in absolute error
at the target location.  The global reference utility for variable $v$ is:
\begin{equation}
  U_v^{\text{global}} =
  \bigl|\hat{y}^{(\text{ablated}_v)} - y^*\bigr| -
  \bigl|\hat{y}^{(\text{full})} - y^*\bigr|
\end{equation}
where $y^*$ is the verification value from the +6h GFS analysis.  A positive
utility $U_v > 0$ indicates that removing variable $v$ degrades the forecast,
confirming its importance.  The attribution-derived variable importance is computed as
$\sum_{\ell} |A_{v,\ell}|$ (summed over all spatial locations $\ell$).
Note that taking absolute values $|A_{v,\ell}|$ before spatial summation
sacrifices the completeness axiom ($\sum_i \mathrm{IG}_i = F(x) - F(\bar{x})$)
but prevents sign-cancellation across locations: a variable with large
positive attributions at some locations and large negative attributions at
others is important, yet its signed sum could be near zero.
We then compute Spearman $\rho$ between the attribution-ranked and
ablation-ranked variable orderings, computed on time-averaged scores.

\paragraph{Spatial validation.}
\label{sec:spatial_ablation}
We define a sparse 468-point European grid (35--70$\degree$N, 10$\degree$W--40$\degree$E)
at $2\degree$ spacing.  At each point, we perturb a local patch
($P \in \{1,3,5\}$, corresponding to $0.25\degree$,
$0.75\degree$, and $1.25\degree$) using one of three perturbation methods:
mean replacement, scale bias (10\%), or additive noise (10\%).
The spatial utility at grid point $g$ is:
\begin{equation}
  U_g^{\text{spatial}} =
  \bigl|\hat{y}^{(\text{perturbed}_g)} - y^*\bigr| -
  \bigl|\hat{y}^{(\text{full})} - y^*\bigr|
\end{equation}
We use $|U_g^{\text{spatial}}|$ rather than signed $U_g$ for two reasons.
First, taking absolute values prevents sign-cancellation when aggregating
across timestamps.  Second, the allocation protocol requires non-negative
payments: a station whose removal improves the forecast (negative $U_g$)
should be handled by the upstream quality filter
(Section~\ref{sec:related_quality}), not by negative rewards.
Signed attributions are used separately for gaming detection, where
directional harm is the relevant signal
(Section~\ref{sec:gaming_methods}).
The attribution spatial importance at grid point $g$ is computed as
$\sum_v |A_{v,g}|$ (summed over common variables).  Spearman $\rho$
is computed between the spatial attribution importance map and the perturbation
sensitivity map, aggregated over timestamps.
\paragraph{Evaluation metrics.}
\label{sec:metrics}
We assess proxy fidelity via Spearman $\rho$ on time-aggregated rankings,
top-$k$ set overlap ($k \in \{1,3,5\}$ global; $\{5,10,20\}$ spatial),
and bootstrap 95\% CIs (10{,}000 resamples): i.i.d.\ bootstrap over the
$n=24$ common variables for global CIs (naturally wide due to small~$n$),
and block bootstrap (${\sim}140$ spatial blocks of ${\sim}4$ grid points)
for spatial CIs.  Temporal multiplicity is
controlled by Benjamini--Hochberg FDR correction within each
configuration.  Full metric definitions are in
Appendix~\ref{sec:appendix_setup_methods}.

\paragraph{Experimental setup.}
\label{sec:experiments}
We evaluate 2 models (FCN, SFNO), 5 European targets (Zurich, London, Berlin,
Madrid, Oslo), and 3 forecast variables (t2m, u10m, msl) on 60 timestamps
(March 2021--December 2022).  Inputs are
$0.25\degree$ GFS analysis fields; the European grid
defined in Section~\ref{sec:spatial_ablation} is used for spatial
evaluation.  The design spans 30 global and 270 spatial
configurations, plus method analyses ($>$400 total).
Detailed setup and compute logistics are in
Appendix~\ref{sec:appendix_setup_methods}.

\section{Results}
\label{sec:results}

\begin{figure*}[tp]
  \centering
  \includegraphics[width=\linewidth,height=0.9\textheight,keepaspectratio]{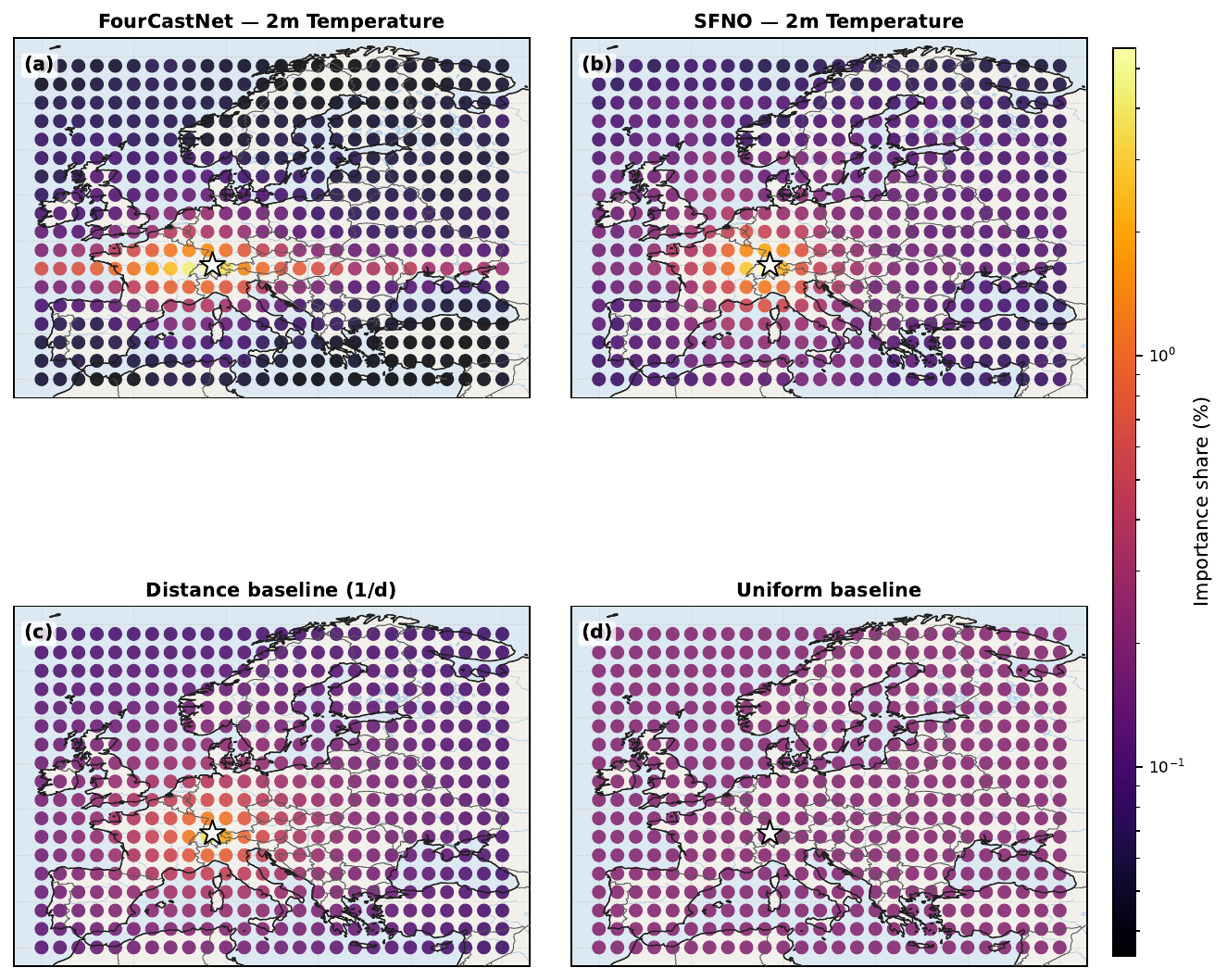}
  \caption{Spatial IG importance maps for Zurich (star marker) for
    2\,m temperature: FourCastNet (a) and SFNO (b), with distance (c)
    and uniform (d) baselines.  Marker colour shows time-averaged
    importance share (\%) on a log scale (fixed marker size in all
    panels).  FCN and SFNO both concentrate mass near the target with
    model-dependent detail; distance is smooth inverse-distance decay;
    uniform is flat.}
  \label{fig:spatial_attribution}
\end{figure*}

Results are organised around five themes: proxy validity, attribution method comparison,
proxy-utility calibration, sensor selection and payment stability, and
gaming resilience.
Extended per-configuration breakdowns and robustness analyses are
provided in the supplementary material.

\subsection{Proxy Validity}
\label{sec:results_validity}

\begin{figure}[t]
  \centering
  \includegraphics[width=\linewidth]{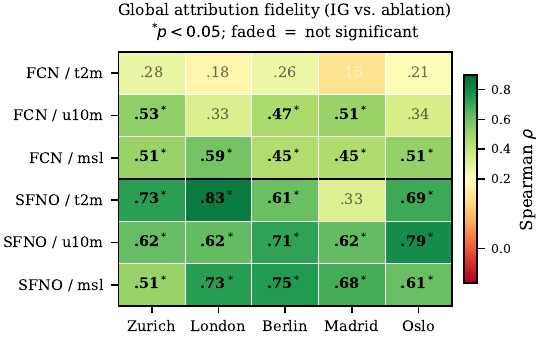}
  \caption{Global attribution fidelity: Spearman $\rho$ between IG
    attribution and ablation utility across 30 configurations
    (2 models $\times$ 5 cities $\times$ 3 target variables).
    Bold values with $^*$ are significant ($p < 0.05$, $n=24$ variables);
    faded values are not.  FCN fails systematically for t2m (all five
    cities non-significant) while SFNO maintains $\rho > 0.5$ in 14
    of 15 configurations.  Per-configuration bootstrap 95\% CIs are
    in the supplementary material (Table~S1).}
  \label{fig:failure_modes}
\end{figure}

\paragraph{Global variable attribution.}
Global variable rankings from IG align with ablation rankings in both
models, with SFNO substantially outperforming FCN (mean
$\rho = 0.655$ vs.\ $0.385$; Figure~\ref{fig:failure_modes}).  Top-5
variable overlap with ablation is high in both models (76\% FCN;
71\% SFNO, Table \ref{tab:supplementary}), indicating that the highest-value variables are recovered
even when full-rank agreement is moderate.  Statistical significance
holds across configurations: Wilcoxon signed-rank tests on per-timestamp $\rho$ values are significant in 14/15
configurations for both models after temporal multiplicity correction
(Table~\ref{tab:methods_summary}).

Attribution fidelity is variable-dependent: pressure and wind are
more attributable than temperature (FCN: msl $\bar{\rho} = 0.502$,
u10m $0.434$, t2m $0.217$; SFNO uniformly high,
$\bar{\rho} \geq 0.639$; Figure~\ref{fig:failure_modes};
supplementary material, Section~S1).
Deployment implications of this variable-dependent fidelity are
discussed in Section~\ref{sec:discussion_deployment}.

\paragraph{Spatial attribution.}
Spatial fidelity is moderate in full-rank terms but strong where
placement decisions matter (Figure~\ref{fig:spatial_attribution}).  For scale perturbations at patch~5,
IG--ablation Spearman~$\rho$ is $0.362$ (FCN) and $0.252$ (SFNO),
while top-5 overlap reaches 72\% and 77\%
(Table~\ref{tab:spatial_fidelity});
block-bootstrap 95\% CIs exclude zero in 26 of 30
scale-perturbation configurations at patch~5;
all four exceptions are temperature configurations,
consistent with the weaker t2m signal noted before.  Correlations increase
monotonically with patch size and are near zero for additive noise,
indicating that structured perturbations are required for informative
spatial validation (per-patch and per-perturbation breakdowns in
supplementary material, Section~S2).

\emph{Cycle-versus-cycle} correlations, each timestamp's attribution
against that timestamp's ablation utility, are low
($\bar{\rho}_t = 0.02$), reflecting volatility in the
single-realisation ablation ground truth rather than
uninformativeness of the attribution map.
\emph{Cycle-versus-aggregate} correlations, each timestamp's
attribution against the time-averaged utility, show that a single
forecast cycle already captures 93\% of the aggregated spatial
fidelity for strong configurations
(Appendix~\ref{sec:appendix_signal_quality}), confirming that the
stable spatial pattern is present in each cycle.

At finer scales, top-5 overlap remains meaningful even in the
single-pixel regime (63--65\% at patch~1), though correlations over all stations
are near detection threshold
($\rho \approx 0.03$--$0.09$; Table~\ref{tab:spatial_fidelity}).
Attribution maps show strong distance decay on average
($\rho=-0.81$ FCN, $-0.92$ SFNO;
supplementary material, Section~S2), establishing proximity as a
strong baseline while leaving room for non-local deviations
relevant to placement.

\paragraph{Model-dependent failure modes.}
Of the 30 global ablation configurations, 9 (30\%) produce confidence intervals
spanning zero ($n=24$ variables limits statistical power) or Spearman
$\rho < 0.2$.  The failure rate is strongly
model-dependent: FCN fails in 8 of 15 configurations (53\%) versus only 1
of 15 for SFNO (7\%, Madrid/t2m only).  Temperature is the most challenging
variable (60\% failure rate); pressure never fails.  However, Wilcoxon
signed-rank tests confirm a systematically positive signal in 14/15
FCN configurations (Table~\ref{tab:methods_summary}), indicating that
the attribution signal is present but too weak for reliable ranking
at $n=24$ variables.  This is consistent with
SFNO's spectral basis enforcing smoothness that is more amenable to
gradient-based attribution, while 2m temperature depends on local surface
processes that create sharper sensitivity patterns
(Figure~\ref{fig:failure_modes}).
A model--task crossover is also apparent: SFNO outperforms FCN for
global variable ranking ($\bar{\rho} = 0.655$ vs.\ $0.385$) but
underperforms spatially ($0.252$ vs.\ $0.362$ at patch~5 scale~10\%),
with SFNO producing negative spatial correlations for t2m at several
cities (e.g., Madrid $\rho = -0.26$, Berlin $\rho = -0.14$).
Deployment implications of this variable-dependent fidelity mismatch
are discussed in Section~\ref{sec:discussion_deployment}.

\paragraph{Baseline sensitivity.}
Zero-baseline IG increases Spearman~$\rho$ by $+0.32$ on average
(range $+0.11$ to $+0.49$; supplementary material, Section~S7), ruling out shared baseline arithmetic
as the coupling mechanism between proxy and ablation; the
climatological-baseline correlations
($\bar\rho = 0.385$--$0.655$; Table~\ref{tab:methods_summary}) are therefore a conservative lower
bound.  A persistence baseline (6-hour prior analysis) produces
uniformly zero attributions, confirming that effective baselines
must be sufficiently distant from the input to traverse informative
gradient regions.  Full per-configuration detail is in the
supplementary material (Section~S7).

\begin{table}[t]
\centering
\caption{Spatial ablation fidelity: mean Spearman $\rho$ and top-5 overlap
  between IG importance ranking and perturbation sensitivity ranking,
  averaged across 15 city--variable configurations.}
\label{tab:spatial_fidelity}
\scriptsize
\begin{tabular}{@{}ll@{\;}c@{\;}c@{\;}c@{\;}c@{\;}c@{\;}c@{}}
\toprule
& & \multicolumn{3}{c}{SFNO} & \multicolumn{3}{c}{FCN} \\
\cmidrule(lr){3-5} \cmidrule(lr){6-8}
Perturbation & Metric & P5 & P3 & P1 & P5 & P3 & P1 \\
\midrule
\multirow{2}{*}{Scale (10\%)}
  & $\rho$  & 0.252 & 0.173 & 0.089 & 0.362 & 0.227 & 0.028 \\
  & Top-5   & 0.77  & 0.79  & 0.65  & 0.72  & 0.76  & 0.63  \\
\multirow{2}{*}{Mean replace}
  & $\rho$  & 0.145 & 0.086 & 0.032 & 0.141 & 0.109 & 0.070 \\
  & Top-5   & 0.80  & 0.68  & 0.47  & 0.76  & 0.69  & 0.48  \\
Noise (10\%)
  & $\rho$  & $-$0.002 & 0.017 & $-$0.015 & 0.012 & 0.002 & 0.038 \\
\bottomrule
\end{tabular}
\end{table}

\subsection{Attribution Method Comparison}
\label{sec:results_methods}

Input-relative scaling is the key property separating useful from
unreliable proxies.  Across all 30 global configurations, VG (no scaling)
is anti-correlated on average ($\bar{\rho}=-0.186$), whereas GTI and IG
are positive ($0.430$ and $0.520$).  IG is significant in 28/30
configurations (Wilcoxon signed-rank on per-timestamp $\rho$),
GTI in 28/30, and VG in only 1/30 (and that case is
significantly negative; Table~\ref{tab:methods_summary}).  Path integration provides a smaller but
consistent gain over GTI: IG exceeds GTI in 24/30 configurations with a
mean advantage of $+0.09$.  GTI retains 83\% of IG's global signal at
one backward pass, making it the better cost--accuracy trade-off when
compute is constrained.  $K=8$ integration steps produce identical
Spearman $\rho$ to $K=50$ in all tested configurations
(supplementary material, Section~S3), confirming rapid convergence.

In the spatial placement task, the method gap disappears: IG/GTI/VG
achieve near-identical oracle ratios ($\sim$94\% at $K=20$) and high
mutual station overlap (91--94\% at top-20).  Spatial aggregation over variables at
each grid point suppresses the variable-scale mismatch that drives VG's
failure in global variable ranking.

The practical implication is task stratification: for \emph{which variables}
drive forecasts, scaled methods (IG or GTI) are necessary; for \emph{where
to place sensors}, even VG suffices.  Method-level detail and
spatial method comparisons are provided in the supplementary material (Section~S3).

\begin{table}[t]
\centering
\caption{Attribution method comparison (summary): mean Spearman $\rho$
  across global ablation configurations, aggregated Spearman significance counts ($p < 0.05$ on time-pooled ranking),
  Wilcoxon signed-rank significance counts ($p < 0.05$ on per-timestamp $\rho$),
  and mean top-5 overlap with ablation ranking.
  Full per-configuration breakdown in
  supplementary material (Section~S3).}
\label{tab:methods_summary}
\resizebox{\columnwidth}{!}{%
\footnotesize
\begin{tabular}{@{}lccccccccccc@{}}
\toprule
& \multicolumn{3}{c}{Mean $\rho$} & \multicolumn{3}{c}{Agg.\ sig.} & \multicolumn{3}{c}{Wilcoxon sig.} & Top-5 \\
\cmidrule(lr){2-4} \cmidrule(lr){5-7} \cmidrule(lr){8-10} \cmidrule(lr){11-11}
Model & IG & GTI & VG & IG & GTI & VG & IG & GTI & VG & IG \\
\midrule
SFNO    & \textbf{0.655} & 0.484 & $-$0.157 & 14/15 & 12/15 & 1/15 & 14/15 & 14/15 & 1/15 & 0.71 \\
FCN     & \textbf{0.385} & 0.375 & $-$0.216 & 8/15  & 6/15  & 0/15 & 14/15 & 14/15 & 0/15 & 0.76 \\
\midrule
Overall & \textbf{0.520} & 0.430 & $-$0.186 & 22/30 & 18/30 & 1/30 & 28/30 & 28/30 & 1/30 & 0.733 \\
\bottomrule
\end{tabular}}
\end{table}

\begin{table*}[t]
\centering
\caption{Proxy family properties.  Backward passes are per inference cycle
  ($K=50$ for IG).  Empirical rows report FCN\,/\,SFNO values.
  Temporal stability values from pairwise timestamp analysis
  (FCN 0.879, SFNO 0.956 for IG; see supplementary Section~S4).
  Global rank calibration from method comparison analysis
  (see supplementary Section~S3).
  Spatial oracle ratio is the mean fraction of oracle utility captured
  at $K=20$ (from Table~\ref{tab:incentive_strategies}).
  Overpayment and Gini ratio from spatial calibration analysis
  (Section~\ref{sec:results_calibration}).}
\label{tab:proxy_properties}
\footnotesize
\begin{tabular}{@{}lcccc@{}}
\toprule
Property & IG & GTI & VG & Distance \\
\midrule
Backward passes & $K{+}1$ & 1 & 1 & 0 \\
Completeness axiom & Yes\textsuperscript{$\dagger$} & No & No & No \\
Input-scale invariance & Yes & Yes & No & N/A \\
\addlinespace
\multicolumn{5}{l}{\textit{FCN\,/\,SFNO}} \\
Temporal stability ($\bar\rho$) & .879\,/\,.956 & .981\,/\,.995 & .994\,/\,.995 & 1.0\,/\,1.0 \\
Global rank calibration ($\bar\rho$) & .385\,/\,.655 & .375\,/\,.484 & $-$.216\,/\,$-$.157 & N/A\textsuperscript{$\ddagger$} \\
Spatial oracle ratio & 93.8\,/\,94.5\% & 94.2\,/\,94.5\% & 93.9\,/\,94.6\% & 91.4\,/\,93.8\% \\
Overpayment (\%) & 33\,/\,36 & 33\,/\,36 & 34\,/\,36 & 55\,/\,47 \\
Gini ratio & .69\,/\,.58 & .73\,/\,.59 & .70\,/\,.56 & .38\,/\,.32 \\
\bottomrule
\end{tabular}
\begin{flushleft}
\textsuperscript{$\dagger$}Completeness holds for signed IG;
unsigned $|\mathrm{IG}|$ (used for reward allocation) sacrifices completeness
for non-negative payment shares.\\
\textsuperscript{$\ddagger$}Distance is purely spatial and cannot rank variables.
\end{flushleft}
\end{table*}

\subsection{Proxy-Utility Calibration and Overpayment}
\label{sec:results_calibration}

\begin{figure*}[t]
  \centering
  \includegraphics[width=\linewidth]{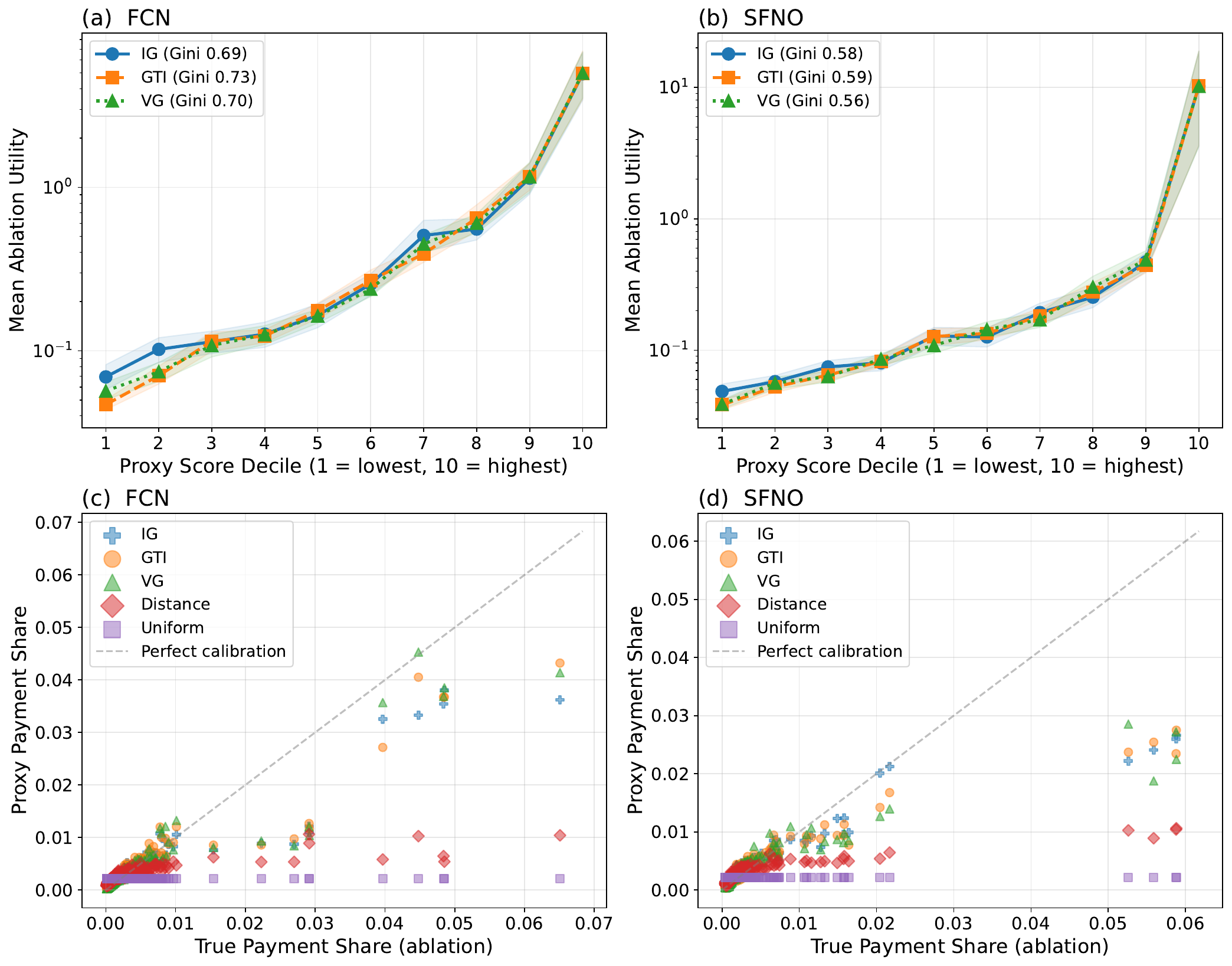}
  \caption{Cardinal calibration and payment accuracy.
    \textbf{(a)--(b)}~Mean ablation utility by proxy score decile
    (log scale): all three gradient methods produce nearly overlapping,
    monotonically increasing curves with sub-unity Gini ratios
    (legend), confirming cardinal fidelity.
    \textbf{(c)--(d)}~Proxy vs.\ true (ablation) payment share for
    each of 468~stations; points above the 45\degree\ line are
    overpaid.  Gradient proxies concentrate near the diagonal;
    distance and uniform allocations show wider scatter.}
  \label{fig:calibration}
\end{figure*}

Proxy scores are cardinally, not just ordinally, faithful to
ablation-based reference utility: binning the 468~stations into deciles
by proxy score yields nearly monotonically increasing mean ablation utility for
all three gradient methods in both models, with nearly overlapping curves
(Figure~\ref{fig:calibration}a--b) and Gini ratios (Gini of proxy
scores divided by Gini of ablation utilities) of
$0.69$/$0.73$/$0.70$ (FCN: IG/GTI/VG) and $0.58$/$0.59$/$0.56$
(SFNO; Table~\ref{tab:proxy_properties}), confirming that method choice barely affects cardinal
calibration.  Ratios below~1 indicate that the proxy
under-concentrates payments relative to true utility: high-value
stations are systematically underpaid while low-value stations are
overpaid.

Beyond aggregate calibration, per-configuration Spearman rank correlations
between proxy and ablation payment shares show that the proxy preserves
station rankings within individual regimes: median $\rho = 0.69$ (FCN) and
$0.50$ (SFNO) across all spatial configurations and gradient methods.

Gradient proxies (IG and GTI) achieve 33--36\% overpayment; distance-based
allocation is substantially worse at 47--55\%; uniform allocation
reaches 61--72\%, roughly double the gradient proxies
(Table~\ref{tab:proxy_properties}; Figure~\ref{fig:calibration}c--d).
Distance Gini ratios (0.38/0.32 FCN/SFNO) are roughly half those of
the gradient proxies (0.56--0.73), confirming that distance spreads
budget far more evenly than true utility warrants.
This gap is universal: distance overpayment exceeds gradient overpayment
in every patch size, perturbation type, and city (per-model breakdowns in
Table~\ref{tab:proxy_properties}).

By budget balance, total underpayment equals total overpayment;
stations at the highest-value locations are systematically underpaid.

\subsection{Sensor Selection and Payment Stability}
\label{sec:results_incentive}

\paragraph{Near-optimal selection.}
Across all tested budgets, gradient methods capture $\geq 92\%$ of
oracle utility.  At $K=20$, any gradient method captures $\sim$62\%
of total utility ($14\times$ uniform efficiency;
Table~\ref{tab:incentive_strategies}).  Distance-based selection also
achieves high efficiency (91--94\% oracle), confirming that
spatial proximity dominates placement value
(cf.\ Figure~\ref{fig:spatial_attribution}c--d); the
attribution-specific advantage (+0.9~pp, 62\% win rate) is
concentrated in FCN (73\%), wind (70\%), and complex terrain
(Zurich: 71\%), settings where the model's sensitivity structure is
more heterogeneous than the smooth distance prior can capture.  For
smoother SFNO/msl settings, distance achieves comparable placement.
These regime-dependent breakdowns are in the
supplementary material (Section~S5).

\paragraph{Payment distribution.}
Under attribution-proportional allocation with a hypothetical budget of
USD~10{,}000/month (Eq.~\ref{eq:payment}), the most important sensor receives
a mean of 14.0\% of the total budget (USD~1{,}397/month), while the median
sensor receives USD~7.88/month (0.08\%)
(supplementary material, Section~S5).  FCN concentrates payments more
sharply (top station: 17.0\%) than SFNO (10.9\%), consistent with FCN's more
localised sensitivity.

\paragraph{Payment stability.}
Bootstrap 95\% CIs on per-station payment shares yield a
CI-to-share ratio of 15.3\% (FCN) and 23.2\% (SFNO) for the top-20
stations (Appendix~\ref{sec:appendix_setup_methods}), meaning payments fluctuate by less than a quarter of their
mean value across forecast cycles.
Global attribution rankings are highly stable across forecast cycles
(pairwise Spearman~$\rho = 0.879$--$0.956$ for IG), while spatial
rankings show moderate stability ($\rho = 0.613$--$0.800$;
supplementary material, Section~S4).

A shrinkage estimator blending the proxy with a distance prior
($\lambda \cdot A_{\text{proxy}} + (1-\lambda) \cdot A_{\text{dist}}$,
fitted by leave-one-timestamp-out cross-validation) confirms that
attribution carries the primary information
($\lambda = 0.67$ FCN, $0.75$ SFNO).  However, $\lambda$ varies
substantially across configurations ($\sigma_\lambda \approx 0.25$),
meaning the blend would not generalise without per-regime tuning.
Since blending does not significantly improve station ranking
(mean $\Delta\rho < 0.004$) and fitting $\lambda$ requires ablation
ground truth, the pure attribution proxy is the recommended default
(full details in Appendix~\ref{sec:appendix_setup_methods}).

\begin{table*}[t]
\centering
\caption{Captured forecast utility (\%) under sensor selection strategies at varying
budget levels $K$ (number of selected stations), averaged across 180 spatial ablation configurations
(2 models $\times$ 5 cities $\times$ 3 variables $\times$ 3 patches $\times$ 2 perturbations).
All three gradient methods achieve $\geq 92\%$ of oracle utility and are statistically
indistinguishable from each other ($p > 0.05$, paired Wilcoxon).
The gradient--distance gap is small ($\sim$1~pp) but highly significant
for $K \geq 10$ (paired Wilcoxon, $p < 10^{-3}$).  Last column reports the mean
IG--distance captured-utility gap (percentage points), with significance from paired
Wilcoxon tests across configurations ($^{***}p<10^{-3}$; ns = not significant).}
\label{tab:incentive_strategies}
\footnotesize
\begin{tabular}{@{}rcccccccc@{}}
\toprule
Budget $K$ & IG & GTI & VG & Distance & Uniform & Oracle & IG/Oracle & IG$-$Distance \\
\midrule
5   & 44.9 & 44.9 & 44.6 & 44.6 & \phantom{0}1.1 & 47.5 & 92.6\% & +0.3 (ns) \\
10  & 53.3 & 53.5 & 53.3 & 52.5 & \phantom{0}2.1 & 56.6 & 92.8\% & +0.8$^{***}$ \\
20  & 61.7 & 61.8 & 61.8 & 60.8 & \phantom{0}4.3 & 65.0 & 94.2\% & +0.9$^{***}$ \\
50  & 72.2 & 72.2 & 72.2 & 71.2 &           10.7 & 75.7 & 94.7\% & +1.0$^{***}$ \\
100 & 80.0 & 80.2 & 80.1 & 79.0 &           21.4 & 84.1 & 94.6\% & +1.0$^{***}$ \\
\bottomrule
\end{tabular}
\end{table*}

\subsection{Gaming Detection: Capabilities and Limits}
\label{sec:results_gaming}

We evaluate anomaly-inflation and data-fabrication attacks across
6 model--city--variable configurations (2~models $\times$ 3~targets;
Section~\ref{sec:gaming_methods}).

\paragraph{Gaming economics: bounded inflation.}
Attackers inflate their mean proxy score by 0.3--4.6\% at
base-design magnitudes (10--50\% perturbation), scaling
monotonically with perturbation magnitude; extended magnitudes
reach higher (supplementary material, Section~S8).
SFNO shows stronger and more
consistent inflation scaling than FCN.  Forecast impact is negligible on average
(MAE increase $< 0.001$ for SFNO), though FCN exhibits rare outlier
increases up to $+38\%$ in coordinated 5-attacker scenarios.
Honest-station collateral damage is small (0.003--0.011~pp per event).

\paragraph{Detection with baseline data.}
The best unsupervised detector, D4~(proxy log-ratio), achieves 100\%
top-5 hit rate on all three SFNO configurations and 58--83\% on FCN,
with PR-AUC of 0.46--0.49 (SFNO) and 0.28--0.36 (FCN;
Table~\ref{tab:gaming_summary}).  Sign-based
and change-point detectors perform at or near chance.  SFNO scenarios
are 1.4--2.3$\times$ more detectable than FCN
(D4: $1.45\times$, D3: $2.34\times$).  A supervised classifier (D7) with
cross-configuration validation does not substantially outperform D4,
suggesting that the detection ceiling is set by signal strength rather
than method sophistication.

\paragraph{Climatological-mean spoofing.}
Climatological-mean spoofing, in which attackers submit long-term
means without deploying physical stations, reduces proxy scores to
$0.94\times$ (FCN) and $0.86\times$ (SFNO) of honest-station levels
(supplementary material, Section~S8).
Because the attacker incurs zero infrastructure cost, the residual
reward remains economically viable.  Inflation detectors (D3, D4)
fail entirely on spoof attacks (top-5 hit rate~0\%), but the
spatial-residual detector (D5) identifies SFNO spoof scenarios with
82.8\% top-5 hit rate by flagging the conspicuous absence of local
anomaly signal.  This asymmetry, inflation detectable but spoofing
not unless spatial structure is monitored, confirms that the
attribution proxy alone does not prevent zero-cost data fabrication;
the staking and identity mechanisms discussed in
Section~\ref{sec:discussion} are necessary.

\paragraph{Baseline-free detection fails.}
All seven baseline-free detectors perform near chance level
(PR-AUC~$\leq 0.028$ vs.\ chance~0.006;
six of seven are $\leq 0.013$, with U6 reaching $0.028$).
Detection requires external reference
information such as historical baselines or quality-control
pipelines~\cite{fenner2021crowdqc}
(Table~\ref{tab:gaming_summary}).

\begin{table}[h]
\centering
\caption{Gaming detection summary across 6 model--city--variable
  configurations (inflate-only scenarios).  Proxy inflation is the mean
  ratio of gaming to baseline proxy scores.  D4 and D3 are the two best
  unsupervised baseline-dependent detectors (proxy log-ratio and
  rank-jump).  D7 shows supervised classifier per-scenario PR-AUC
  (logistic regression, LOCO cross-validation).  Top-5 hit rate is
  for D4.  Zurich/t2m uses 1{,}230 inflate-only scenarios (uniform
  + stratified); other configurations use 270 uniform scenarios each.}
\label{tab:gaming_summary}
\scriptsize
\begin{tabular}{@{}ll@{\;}c@{\;}c@{\;}c@{\;}c@{\;}c@{}}
\toprule
Model & Target & Inflation & D4 PR-AUC & D3 PR-AUC & D4 top-5 & D7 (LR) \\
\midrule
FCN  & Zurich/t2m  & 1.024 & 0.276 & 0.118 & 0.58 & 0.20 \\
FCN  & Zurich/u10m & 1.011 & 0.343 & 0.239 & 0.83 & 0.17 \\
FCN  & London/t2m  & 1.010 & 0.355 & 0.251 & 0.79 & 0.17 \\
SFNO & Zurich/t2m  & 1.063 & 0.488 & 0.468 & 1.00 & 0.43 \\
SFNO & Zurich/u10m & 1.017 & 0.460 & 0.481 & 1.00 & 0.28 \\
SFNO & London/t2m  & 1.024 & 0.465 & 0.474 & 1.00 & 0.32 \\
\bottomrule
\end{tabular}
\end{table}

\section{Discussion}
\label{sec:discussion}

\subsection{Deployment Architecture and Mechanism Design Implications}
\label{sec:discussion_deployment}

Gradient-based attribution fills a missing layer in participatory
weather sensing, a model-derived value signal for reward allocation.
Existing incentive mechanism designs assume such a signal as
input~\cite{yang2012crowdsourcing,zhang2016incentives}, yet deployed
networks implement quality gating~\cite{fenner2021crowdqc} and, in
some cases, hardware-based Sybil resistance~\cite{chiu2025depin},
but allocate rewards by quality or geometry rather than predictive
contribution.  For instance,
WeatherXM~\cite{weatherxm2024rewards} distributes payments
proportional to data quality and uptime, not downstream utility
(Section~\ref{sec:related_quality}).
The signal characterised here closes this gap: gradient proxies
achieve 33--36\% overpayment versus 61--72\% for quality-gated
uniform allocation, roughly halving
budget misallocation.  Moreover, because each forecast query
produces a different backward-pass attribution map, the mechanism
enables \emph{usage-driven} reward allocation that directs network
growth toward demand rather than distributing rewards uniformly
across quality-passing stations.
However, attribution alone is not sufficient: climatological-mean
spoofing evades all inflation detectors while retaining 86--94\% of
honest-station rewards (Section~\ref{sec:results_gaming}), so
staking~\cite{kraner2023agent} or identity mechanisms that raise
fabrication cost remain necessary complements.  Systematic
overpayment under current uniform schemes can erode contributor
incentives and threaten long-run network
stability~\cite{kalabic2023burn}; the attribution signal mitigates
this by concentrating payments on high-value stations.

\paragraph{Attribution vs.\ distance: implications for mechanism design.}
The attribution advantage over distance separates into three distinct
value propositions: calibrated payment (robust across configurations),
spatial placement (configuration-dependent), and variable
ranking (not provided by distance-based methods).
The overpayment gap between gradient proxies and distance
(Section~\ref{sec:results_calibration}) holds across every patch size,
perturbation type, and city, a calibration advantage consistent across
all tested configurations.
The \emph{spatial placement} advantage, by contrast, is concentrated in
wind variables, complex terrain, and FCN, i.e., settings where the model's
sensitivity structure is more heterogeneous than the smooth distance
prior can capture.
The observed patch-size interaction (FCN$\times$patch~1: 77\% wins,
$+2.3$\,pp; Table~\ref{tab:supplementary}) is consistent with the hypothesis that finer spatial
resolution produces more heterogeneous sensitivity structures that
distance-based proxies cannot capture.
Whether this translates to higher-resolution model grids, where
the resolved physics differ qualitatively, remains an open empirical
question.
Distance is purely geometric and by construction cannot rank
input variables; attribution uniquely enables variable-level budget
discrimination.
In practice the two tasks should be distinguished: for spatial placement
alone in smooth settings, distance achieves near-equivalent selection;
for calibrated payments, gradient proxies retain a universal advantage.
For networks targeting diverse forecast products or complex terrain,
attribution justifies the additional compute cost for both placement and
payment.
A further deployment consideration is model selection.  The two
architectures exhibit complementary strengths: SFNO dominates global
variable ranking ($\bar\rho = 0.655$ vs.\ $0.385$) while FCN dominates
spatial placement ($0.362$ vs.\ $0.252$ at patch~5;
Section~\ref{sec:results_validity}); FCN also shows stable
out-of-sample performance ($\bar\rho_{\text{OOS}} = 0.401$ vs.\
$0.385$ in-sample, within sampling variability) whereas SFNO degrades
moderately ($0.556$ vs.\ $0.655$; Table~\ref{tab:temporal_oos}).
A deployment targeting both variable-level budget allocation and
spatial placement may therefore benefit from using different models
for each task, selecting SFNO for variable ranking and FCN for
station placement.  Whether this complementarity is a stable
property of the two architectures or specific to our evaluation
setting remains open.

\paragraph{Temporal aggregation for payment computation.}
Single-cycle attributions are weakly correlated with single-cycle
ablation utility ($\bar\rho_t = 0.02$;
Section~\ref{sec:results_validity}), partly reflecting noise in the
single-realisation ablation ground truth rather than uninformativeness
of the attribution map.  In particular, each cycle recovers 93\% of the
fully aggregated spatial fidelity $\rho(\bar{a}, \bar{u})$ for strong
configurations, and Wilcoxon significance emerges at a median of
$N = 12$ timestamps for scale perturbations at patch~5
(Appendix~\ref{sec:appendix_signal_quality}).
Aggregating on the order of 10 to 15 forecast cycles appears
sufficient to stabilise spatial fidelity in strong-signal regimes,
though this estimate is based on sparsely sampled timestamps rather
than consecutive 6-hourly cycles
(Appendix~\ref{sec:appendix_setup_methods}), so the correspondence
to a rolling-window deployment remains an open question.
Convergence speed is regime-dependent: temperature converges more
slowly (median $N = 18.5$) and weaker-signal configurations (noise
perturbation, patch~1) may not converge at all
(Appendix~\ref{sec:appendix_signal_quality}).

\paragraph{Payment concentration.}
All proxy Gini ratios fall below~1 (0.32 to 0.73;
Section~\ref{sec:results_calibration}), meaning proxy payments are
\emph{less unequal} than true utility would warrant, a property that
is either a distributional feature reducing payment inequality or a
mechanism limitation providing insufficient incentive differentiation,
depending on the operator's participation objectives.
Separately, because attribution is computed for a single forecast
target, rewards cluster geographically near it (top station: 14 to
17\% of budget; median: 0.08\%;
Section~\ref{sec:results_incentive}); averaging attribution maps
over diverse forecast targets would spread rewards more broadly but
is not evaluated here.

\paragraph{Subadditivity and redundancy.}
The preceding analysis treats stations independently, but nearby
stations provide partially redundant information: joint ablation
yields less impact than the sum of individual impacts
(subadditivity ratios: FCN 0.75--0.80, SFNO near-additive at 0.96;
supplementary material, Section~S7-E).  A deployed mechanism would
therefore require spatial diversity constraints or
marginal-contribution adjustments.
The absolute captured utility at $K = 20$ (62\%) should accordingly
be understood as an upper bound: accounting for FCN's subadditivity
deflates it to approximately 46\%, while SFNO's near-additive ratio
makes this correction negligible.  The oracle ratio ($\geq 92\%$)
is approximately preserved because redundancy deflates both the
selected set and the oracle set similarly.

\paragraph{Temperature attribution as a deployment constraint.}
Attribution fidelity depends on the \emph{forecast target}, not on
what stations measure: msl ($\bar\rho = 0.502$/$0.655$ FCN/SFNO)
and wind ($0.434$/$0.673$) targets yield reliable proxy signals,
while a t2m target has a 60\% global failure rate, with FCN failing
in all five cities (Section~\ref{sec:results_validity}).
Wind and pressure forecast targets, where the proxy is most reliable,
are therefore the natural entry point for deployment.
The principal mitigation for t2m targets is model selection:
SFNO maintains $\bar\rho = 0.639$ for t2m globally (the minimum across
target variables; supplementary material, Section~S1), largely
resolving the failure mode specific to FCN.

\subsection{What the Proxy Captures}
\label{sec:discussion_epistemic}

\paragraph{From quality assessment to valuation.}
Quality assessment determines station \emph{eligibility}; our work
addresses \emph{valuation}: how much eligible data is worth for
forecasting.  Real citizen weather stations introduce heterogeneities
(irregular reporting, warm biases~\cite{chapman2017crowdsourcing}) absent
from our gridded evaluation, so operational deployment requires an
upstream quality pipeline (e.g.,
CrowdQC+~\cite{fenner2021crowdqc}) to transform raw reports into
model-compatible fields before the attribution proxy can assign value.

\paragraph{From gridded evaluation to station deployment.}
The AI models ingest gridded analysis fields, not raw station
observations.  The chain from station to forecast spans several
links: a station observation enters data assimilation, producing
an analysis field that the AI model ingests and differentiates.
This paper characterises the last link, namely model sensitivity to its
gridded inputs, under controlled conditions.  Whether a single
station materially changes the analysis at its grid point (an
assimilation-impact question) requires separate validation; together
with the data-quality issues noted in the preceding paragraph,
this motivates field deployment with live data denial as the priority
extension.

\paragraph{Proxy-ablation structural coupling.}
Both the gradient proxy and the ablation benchmark probe model input
sensitivity; their agreement validates the model's
sensitivity structure.  Adjoint-based FSOI faces the same structural
coupling yet was adopted for operational network
optimisation~\cite{langland2004estimation} because it serves the target
task.  Three lines of evidence establish that the agreement is
substantive rather than artifactual:
(i)~decoupling the IG baseline from the ablation replacement value
\emph{increases} $\rho$ by $+0.32$
(Section~\ref{sec:results_validity}), ruling out shared baseline
arithmetic;
(ii)~proxy rankings generalise out-of-sample with comparable fidelity
in both models (Table~\ref{tab:temporal_oos}), ruling out same-sample
overfitting;
(iii)~the $\geq 92\%$ oracle utility result evaluates a station
placement \emph{decision}, not a rank correlation: the proxy
identifies a near-optimal station set under the model's own
utility criterion.

\begin{table}[t]
\centering
\caption{Temporal out-of-sample validation: Spearman $\rho$ between train-set attribution ranking and test-set ablation utility ranking (chronological 30/30 split). Full-sample $\rho$ shown for comparison.}
\label{tab:temporal_oos}
\footnotesize
\begin{tabular}{lcccccc}
\toprule
& \multicolumn{2}{c}{IG} & \multicolumn{2}{c}{GTI} & \multicolumn{2}{c}{VG} \\
\cmidrule(lr){2-3} \cmidrule(lr){4-5} \cmidrule(lr){6-7}
Model & OOS $\rho$ & Full $\rho$ & OOS $\rho$ & Full $\rho$ & OOS $\rho$ & Full $\rho$ \\
\midrule
FCN  & 0.401 & 0.385 & 0.379 & 0.375 & $-$0.238 & $-$0.216 \\
     & \emph{(8/15)} & & \emph{(7/15)} & & \emph{(1/15)} & \\[2pt]
SFNO & 0.556 & 0.655 & 0.438 & 0.484 & $-$0.143 & $-$0.157 \\
     & \emph{(12/15)} & & \emph{(11/15)} & & \emph{(1/15)} & \\
\bottomrule
\end{tabular}\\
\raggedright\footnotesize Parenthetical: number of 15 OOS city--variable configurations where $\rho > 0$.
\end{table}

\subsection{Limitations and Gaming Resilience}
\label{sec:discussion_limitations}

\paragraph{Gaming resilience.}
The proxy is not fabrication-resistant on its own.
Anomaly inflation is detectable with baseline proxy-monitoring
(Section~\ref{sec:results_gaming}), but baseline-free methods fail
entirely.  Climatological-mean spoofing evades inflation detectors
while retaining 86 to 94\% of honest-station rewards at zero
infrastructure cost; only the spatial-residual detector flags it
reliably for SFNO.  A deployed mechanism should therefore assume convergence on data
fabrication and employ complementary safeguards, such as external
quality control~\cite{fenner2021crowdqc} and hardware-based identity
verification or economic staking~\cite{chiu2025depin}, given that
spoofers retain 86 to 94\% of honest-station rewards.

\paragraph{Limitations.}
The study is limited to +6h forecasts, two architectures, and European
targets.
The cost--benefit analysis uses simulated budgets; deployment
requires additional constraints including installation logistics,
maintenance, data transmission, and regulation.  The allocation
analysis assumes static value maps, while operational deployment
may require time-varying updates by forecast regime.
Both models are from NVIDIA's Earth-2 framework, selected for
open-weight differentiability; the substantially different fidelity
between the vision-transformer (FCN, $\bar\rho = 0.385$) and
spectral-operator (SFNO, $0.655$) architectures, together with a
1.4 to 2.3$\times$ gaming detectability gap, indicate that results
are driven by architectural differences rather than shared framework
origin.
Beyond model choice, collusion, temporal mimicry, and adversarial input
crafting are not covered.  Only one data-fabrication variant
(climatological-mean spoofing) was tested; more sophisticated strategies
(synthetic plausible weather, adversarial perturbations toward
high-attribution patterns) remain open.
Detection metrics assume known attack prevalence and independent
timestamps; violations would affect threshold calibration and
Type~I error rates.

\section{Conclusion}
\label{sec:conclusion}

Gradient-based attribution from differentiable AI weather models
provides a cardinally calibrated, computationally feasible value
signal for participatory sensing reward allocation.
We evaluated fidelity, calibration, cost, and gaming resilience
across more than 400~configurations on gridded GFS analysis inputs.

\paragraph{What works.}
Input-relative scaling is the critical property: vanilla gradients
are anti-correlated with ablation utility ($\bar\rho = -0.19$),
while Gradient~$\times$~Input and Integrated Gradients yield
faithful rankings ($\bar\rho = 0.43$--$0.52$).  The cheapest
single-pass variant (GTI) retains 83\% of IG fidelity at
$1/50$th the computational cost, making per-cycle attribution
operationally feasible.  Proxy scores are not merely ordinally but
\emph{cardinally} faithful: payments are monotonically calibrated
against ablation utility (Gini ratios 0.56--0.73), with total
overpayment of 33--36\%, roughly half the 61--72\% incurred by
quality-gated uniform allocation as deployed by existing
participatory weather sensing networks
(Section~\ref{sec:related_quality}).
Payment shares are stable across forecast cycles (bootstrap
CI-to-share ratio 15--23\% for top-20 stations).

\paragraph{Where it matters most.}
Attribution captures $\geq 92\%$ of oracle sensor-placement utility,
with a regime-conditional advantage over distance in complex terrain
and wind forecasting (+1.4~pp, 73\% win rate).  The tasks separate
cleanly: identifying \emph{which variables} drive forecasts requires
input-scaled methods; identifying \emph{where to place sensors} does
not, as spatial aggregation over variables suppresses the scale
mismatch.  An operator targeting placement alone can therefore use
Vanilla Gradients at single-pass cost.
Model selection further stratifies: SFNO dominates variable ranking
while FCN dominates spatial placement and generalises more stably
out-of-sample, suggesting that deployment may benefit from
task-specific model selection.
For pressure and wind targets at coarser patches, each single
forecast cycle already recovers 93\% of time-averaged spatial
fidelity, enabling per-cycle reward computation; temperature targets
exhibit weaker signals that may not converge regardless of
aggregation length.

\paragraph{What fails.}
The proxy's attack surface has two distinct vulnerabilities.
Anomaly inflation is detectable with baseline proxy-monitoring
(100\% top-5 hit rate, SFNO), but baseline-free detection fails
entirely (PR-AUC~$\leq 0.028$ vs.\ chance 0.006): detection
requires external reference information.
Separately, climatological-mean spoofing evades all inflation
detectors while retaining 86--94\% of honest-station rewards at zero
infrastructure cost.  The attribution signal does not prevent data
fabrication; staking-based and identity mechanisms remain necessary
complements.

These results establish gradient attribution as the first
computationally validated candidate signal for model-informed reward
allocation in participatory weather sensing.  Whether model input
sensitivity faithfully reflects observation value must be established
through field deployment, closing the station-to-grid link that
this study deliberately isolates, alongside longer lead times and
broader geographic regimes.  The attribution advantage over distance
is most pronounced at the finest resolution tested
(FCN at patch~1: 77\% win rate, $+2.3$\,pp;
Table~\ref{tab:supplementary}), motivating evaluation on
higher-resolution model grids where richer sensitivity structure may
further widen this gap.  Nevertheless, the signal's cardinal
faithfulness, its variable-identification capability unique among
proxies, and its computability at forecast time establish it as a
validated candidate component for multi-component incentive mechanisms.  More broadly, the underlying
approach, differentiating a differentiable model's prediction with
respect to its inputs to derive a value signal, is not specific to
weather: any participatory sensing domain whose data feeds a
differentiable model can derive value signals in the same way.

\appendices

\section{Experimental Setup and Extended Methods}
\label{sec:appendix_setup_methods}

\paragraph{Targets and variables.}
We evaluate 5 cities with fixed target coordinates: Zurich
(47.4$\degree$N, 8.6$\degree$E), London (51.5$\degree$N, 0.1$\degree$W),
Berlin (52.5$\degree$N, 13.4$\degree$E), Madrid (40.4$\degree$N, 3.7$\degree$W),
and Oslo (59.9$\degree$N, 10.8$\degree$E).  For each city, we evaluate
3 target variables (t2m, u10m, msl), yielding 15 city--variable targets
per model.

\paragraph{Timestamps.}
We use 60 timestamps sampled from March 2021 to December 2022 (12 DJF, 16 MAM,
17 JJA, 15 SON; initialisation hours 00/06/12/18 UTC).

\paragraph{Input domain and spatial crop.}
Model inference and attribution run on each model's native grid using GFS
analysis inputs at $0.25\degree$ resolution.  The Europe crop
(35--70$\degree$N, 10$\degree$W--40$\degree$E) is introduced for spatial
perturbation experiments and sensor-selection evaluation only, using a sparse
candidate grid of 468 points at $2\degree$ spacing ($18\times26$).

\paragraph{Configuration counts.}
The core experiment includes 30 global configurations
($2$ models $\times$ $5$ cities $\times$ $3$ target variables) and
270 spatial configurations
($2$ models $\times$ $5$ cities $\times$ $3$ target variables
$\times$ $3$ patch sizes $\times$ $3$ perturbation modes).  Additional method-comparison and robustness
experiments (e.g., $K$-sensitivity, baseline sensitivity, per-variable spatial
ablation) bring the total to more than 400 configurations.

\paragraph{Compute.}
Runs were executed on UZH ScienceCluster GPUs (A100/H100/L4).  Total project
compute was approximately 920 GPU-hours, of which 380 for the main attribution
pipeline (${\sim}$120 SLURM jobs), 340 for gaming experiments, and 180 for
revision-phase robustness analyses, plus CPU post-processing.

\paragraph{Aggregation convention.}
Reported mean Spearman $\rho$ values are unweighted means across configurations.
Each configuration-level $\rho$ is computed on time-pooled rankings from the
60 timestamps.

Table~\ref{tab:supplementary} collects key derived quantities cited in the
main text, with their sample sizes, to facilitate reproducibility.

\begin{table}[h]
\centering
\caption{Supplementary statistics referenced in the main text.  All values
  are computed from time-pooled rankings and averaged across the stated
  configurations.}
\label{tab:supplementary}
\footnotesize
\begin{tabular}{@{}llcc@{}}
\toprule
Statistic & Scope & Value & $n$ \\
\midrule
\multicolumn{4}{l}{\emph{Global ablation: top-$k$ overlap (IG vs.\ ablation)}} \\
FCN mean top-1 overlap  & 15 configs & 47\% & 15 \\
SFNO mean top-1 overlap & 15 configs & 20\% & 15 \\
FCN mean top-5 overlap  & 15 configs & 76\% & 15 \\
SFNO mean top-5 overlap & 15 configs & 71\% & 15 \\
\midrule
\multicolumn{4}{l}{\emph{Spatial: method station overlap}} \\
IG--GTI top-20 overlap  & 180 configs & 94\% & 180 \\
IG--VG top-20 overlap   & 180 configs & 91\% & 180 \\
\midrule
\multicolumn{4}{l}{\emph{Spatial: IG vs.\ distance vs.\ oracle (top-5)}} \\
IG--distance overlap    & 180 configs & 77\% & 180 \\
IG--oracle overlap      & 180 configs & 73\% & 180 \\
Distance--oracle overlap & 180 configs & 71\% & 180 \\
\midrule
\multicolumn{4}{l}{\emph{IG vs.\ distance: conditional win rates}} \\
FCN (all patches)       & 450 comparisons & 73\%, $+1.4$~pp & 450 \\
SFNO (all patches)      & 450 comparisons & 52\%, $+0.2$~pp & 450 \\
Wind (u10m)             & 300 comparisons & 70\%, $+1.2$~pp & 300 \\
Zurich                  & 180 comparisons & 71\%, $+1.2$~pp & 180 \\
Patch 1                 & 300 comparisons & 66\%, $+1.2$~pp & 300 \\
Patch 3                 & 300 comparisons & 59\%, $+0.6$~pp & 300 \\
Patch 5                 & 300 comparisons & 62\%, $+0.6$~pp & 300 \\
FCN $\times$ Patch 1    & 150 comparisons & 77\%, $+2.3$~pp & 150 \\
\bottomrule
\end{tabular}
\end{table}

\paragraph{Block bootstrap details.}
Block bootstrap CIs use ${\sim}140$ spatial blocks of 4 grid points each
($2\times 2$ at $4\degree$ spacing), preserving local autocorrelation within
each resampled block.  Compared to standard (i.i.d.)\ bootstrap, block CIs
are 22\% wider at patch~1 (mean inflation $1.22\times$, effective
$n_{\text{eff}} = 355$) and 40\% wider at patch~5 (inflation $1.40\times$,
$n_{\text{eff}} = 251$), reflecting moderate spatial dependence.  In 107 of
120 configurations, the block CI is wider than or equal to the standard CI.

\paragraph{Assumptions.}
(a)~Stations report time-stamped meteorological fields to a central
coordinator who runs the AI weather model and has a fixed reward
budget~$B$ per forecast window.
(b)~Ablation audits require verification data and are feasible offline
but not at forecast time, motivating the need for a proxy computable
from the prediction alone. Additionally, ablation scales $O(N)$ with
the number of candidate stations (one additional forward pass per
station per forecast target), whereas the gradient proxy requires a
single backward pass per target regardless of~$N$.
(c)~Attackers inflate anomalies at their own station(s); we do not model
collusion across stations or adversarial input crafting that targets
model vulnerabilities directly.
(d)~The ablation-based utility used for validation measures single-station
removal impact, the operationally relevant counterfactual for a running
network (what forecast quality is lost if station~$g$ goes offline).
For axiomatic fair division, Shapley values are the theoretically correct
reference but are computationally infeasible at operational scale
(supplementary material, Section~S7-D) and average over coalition
configurations (e.g., networks with fewer than 10 of 468~stations) that
do not arise operationally.  The two measures diverge under redundancy;
the gap is indicated by the subadditivity ratios in
Section~\ref{sec:discussion_deployment} (SFNO: 0.96, negligible; FCN: 0.75--0.80,
implying that redundant stations are over-credited by single-station
ablation, motivating the spatial diversity constraints recommended there).

\paragraph{Calibration definitions.}
The Gini ratio (Gini of proxy scores divided by Gini of ablation
utilities) summarises concentration mismatch.  Decile calibration curves
bin the 468~stations by proxy score and plot mean ablation utility per
decile.  Symmetrically, \emph{underpayment} at station $g$ is
$\max(0,\, p_{\text{true}}(g) - p_{\text{proxy}}(g))$; total underpayment
equals total overpayment by budget balance.

\paragraph{Evaluation metric definitions.}
Spearman $\rho$ is computed on time-aggregated rankings.
Top-$k$ overlap uses $k \in \{1,3,5\}$ for global ablation and
$\{5,10,20\}$ for spatial ablation.
All reported spatial CIs use block bootstrap (${\sim}140$ blocks of 4
grid points at $4\degree$ spacing).
Benjamini--Hochberg FDR correction is applied within each configuration
to $n = 60$ per-timestamp Spearman $p$-values.
Wilcoxon signed-rank tests assess whether per-timestamp $\rho$ values are
systematically positive.

\paragraph{Payment uncertainty.}
Per-station proxy scores vary across timestamps.  We use i.i.d.\ bootstrap
(resampling individual timestamps with replacement); because the 60
timestamps are sparsely sampled across seasons and years with irregular
gaps, temporal block structure does not preserve meaningful autocorrelation.
The shrinkage estimator is
$A_{\text{shrunk}}(g) = \lambda \cdot A_{\text{proxy}}(g) +
(1 - \lambda) \cdot A_{\text{dist}}(g)$, where $\lambda$ is fitted by
leave-one-timestamp-out cross-validation minimising MSE against ablation
utility.  Shrinkage can change station rankings, not just stabilise
shares.
Replacing the MSE inner-loop objective with captured utility at
$k \in \{5, 20\}$ shifts $\lambda$ downward
(FCN: $0.43$/$0.50$; SFNO: $0.23$/$0.52$ vs.\ $0.67$/$0.75$ under MSE)
but does not improve outer-loop ranking quality
($\Delta\rho < 0.004$, $p \ge 0.11$ for both models and objectives),
confirming that no convex blend of proxy and distance improves station
ranking regardless of training criterion.

\section{Signal Quality and Sensitivity Analysis}
\label{sec:appendix_signal_quality}

As discussed in Sections~\ref{sec:results_validity}
and~\ref{sec:discussion_deployment}, each single-timestamp attribution
recovers 93\% of the fully aggregated $\rho(\bar{a}, \bar{u})$ for strong
configurations, and Wilcoxon significance emerges at a median of $N = 12$
timestamps.  This section provides the per-variable and per-perturbation
breakdowns underlying those headline numbers.

\paragraph{Per-variable recovery.}
$\rho(a_t, \bar{u})$ averages $0.10$ across all configurations and $0.29$ for
strong configurations (patch~5, scale~10\%).  Of individual timestamps in the
strong regime, 89\% are individually significant ($p < 0.05$,
Benjamini--Hochberg corrected).  The recovery ratio varies by target variable:
u10m retains 95\% (median), msl and t2m each retain 90\%.

\paragraph{Per-variable and per-mode convergence.}
Among scale perturbation configs at patch~5, 69\% reach Wilcoxon significance.
By variable: msl converges fastest (51\% of configs significant, median
$N = 13$), followed by u10m (43\%, $N = 12$), while t2m is slowest (34\%,
$N = 18.5$).  Configs that never reach significance are predominantly noise
perturbation (29\% significant) and patch~1 (29\%), where the effect size is
near zero rather than merely slow to converge.

\section*{Conflict of Interest}
M.~C.~Ballandies and M.~T.~C.~Chiu are co-founders and board members of
WiHi, an association developing a weather intelligence platform.  The research
question addressed in this paper emerged in part from WiHi's work on incentive
design for participatory weather sensing.  WiHi has not issued tokens or
equity; the authors may receive allocations if such instruments are issued.
C.~J.~Tessone declares no conflict of interest.

\bibliographystyle{IEEEtran}
\bibliography{references}

\end{document}